\documentclass[journal]{IEEEtran}

\usepackage{cite}

\usepackage{amsmath}
\usepackage{tikz}
\usetikzlibrary{positioning}

\usepackage{url}
\usepackage{graphicx}  
\usepackage{float}  

\begin{document}

\title{Explainable AI in Big Data Fraud Detection}

\author{
\begin{tabular}{ccc}
Ayush Jain & Rahul Kulkarni & Siyi Lin \\
\textit{Northeastern University} & \textit{Northeastern University} & \textit{Northeastern University} \\
Boston, USA & Boston, USA & Boston, USA \\
\texttt{jain.a1@northeastern.edu} &
\texttt{kulkarni.rahu@northeastern.edu} &
\texttt{lin.siyi@northeastern.edu}
\end{tabular}
}


\maketitle

\begin{abstract}
Big Data has become central to modern applications in finance, insurance, and cybersecurity, enabling machine learning systems to perform large-scale risk assessments and fraud detection. However, the increasing dependence on automated analytics introduces important concerns about transparency, regulatory compliance, and trust. This paper examines how explainable artificial intelligence (XAI) can be integrated into Big Data analytics pipelines for fraud detection and risk management. We review key Big Data characteristics and survey major analytical tools, including distributed storage systems, streaming platforms, and advanced fraud detection models such as anomaly detectors, graph-based approaches, and ensemble classifiers. We also present a structured review of widely used XAI methods, including LIME, SHAP, counterfactual explanations, and attention mechanisms, and analyze their strengths and limitations when deployed at scale. Based on these findings, we identify key research gaps related to scalability, real-time processing, and explainability for graph and temporal models. To address these challenges, we outline a conceptual framework that integrates scalable Big Data infrastructure with context-aware explanation mechanisms and human feedback. The paper concludes with open research directions in scalable XAI, privacy-aware explanations, and standardized evaluation methods for explainable fraud detection systems.
\end{abstract}


\section{Introduction}

\IEEEPARstart{T}{he} rapid growth of Big Data in sectors such as finance, insurance, and cybersecurity has reshaped how organizations detect fraud, assess risk, and make high-stakes decisions [1]. With technologies like online banking, IoT devices, mobile payments, and interconnected financial systems, massive amounts of diverse data are generated every second. To handle this scale, institutions increasingly rely on advanced machine learning (ML) and artificial intelligence (AI) systems. These models outperform traditional rule-based systems by detecting subtle patterns, adapting to evolving fraud behaviors, and identifying anomalies that humans may overlook [3].

Despite their effectiveness, these automated systems raise important concerns around transparency, accountability, and trust. Many high-performing ML models, such as deep neural networks, ensemble methods, and graph-based fraud detectors, operate as ``black boxes,'' providing little insight into how decisions are made [4]. In sensitive domains such as credit scoring, insurance underwriting, and fraud detection, this lack of interpretability complicates regulatory compliance, internal auditing, and user confidence.

Explainable Artificial Intelligence (XAI) has emerged as a promising way to address these challenges. XAI techniques enhance the interpretability of complex ML models while maintaining strong predictive performance. This becomes especially important in Big Data environments, where models operate across distributed infrastructures, real-time data streams, and high-dimensional feature spaces.

This paper examines how XAI can be integrated into Big Data analytics for risk management and fraud detection. The central research question guiding this work is:

\begin{quote}
\textit{``How can explainable artificial intelligence (XAI) be integrated into Big Data analytics for risk management and fraud detection to improve transparency, regulatory compliance, and trust in automated decision-making systems?''}
\end{quote}

This paper makes three contributions:
\begin{enumerate}
    \item a literature review of Big Data--driven fraud detection and XAI techniques;
    \item a critical analysis of the challenges and limitations involved in integrating XAI into large-scale analytics pipelines; and
    \item the proposal of a conceptual framework for scalable, real-time, explainable fraud detection architectures.
\end{enumerate}

\section{Literature Review}

\subsection{Characteristics of Big Data}
Big Data refers to datasets that are so large, fast moving, or diverse that traditional data management tools cannot handle them effectively. It is commonly described using the ``Four V’s.'' \textit{Volume} represents the massive scale of modern datasets, which may reach terabytes or even petabytes of information collected from sensors, online platforms, and enterprise systems. \textit{Velocity} refers to the speed at which data is generated and the rate at which it must be processed, especially in real-time environments such as financial transactions, cybersecurity monitoring, and Internet of Things (IoT) systems. \textit{Variety} highlights the wide range of formats organizations must handle, including structured databases, semi-structured logs, and unstructured text, images, audio, and video. \textit{Veracity} captures the uncertainty and noise that naturally appear in large real-world datasets. These characteristics explain why traditional relational databases struggle and why modern Big Data systems rely on scalable, distributed tools and architectures [2].

\subsection{Big Data Analytics Tools and Methods}
The growth of Big Data has driven the development of specialized tools for data storage, processing, and analysis. Traditional relational databases are limited in their ability to manage heterogeneous and large-scale data, which has encouraged adoption of more flexible and distributed frameworks. Storage technologies such as the Hadoop Distributed File System (HDFS) distribute data across nodes with replication for reliability. NoSQL databases enable schema-less and horizontally scalable architectures, while in-memory databases eliminate disk I/O bottlenecks and support real-time computation. Analytical frameworks such as MapReduce and Massively Parallel Processing (MPP) allow organizations to perform large computations across distributed clusters. Real-time processing engines, including Apache Spark Streaming and Apache Kafka, support high-velocity pipelines with low latency. Common analytical methods include clustering, classification, regression, association rule mining, text mining, sentiment analysis, and social network analysis. Together, these tools form the foundation of modern Big Data analytics ecosystems [2].

\subsection{Big Data Analytics in Risk and Fraud Detection}
Traditional fraud detection systems relied on rule-based techniques such as threshold checks, blacklists, and predefined anomaly patterns. Although these methods were easy to interpret, they lacked flexibility and struggled to detect new or evolving fraud behavior. Modern fraud detection systems use machine learning and statistical modeling to identify subtle or previously unseen patterns. Big Data plays a central role in enabling these approaches, since fraud detection involves high-volume transactions, real-time streaming data, diverse data formats, and noisy or incomplete inputs. State-of-the-art models include unsupervised anomaly detectors, graph neural networks for identifying fraud rings, temporal models that capture behavioral deviations, and ensemble classifiers trained on large transaction histories. These systems provide strong predictive performance but introduce challenges for transparency, auditability, and regulatory compliance [3][4].

\subsection{Explainable Artificial Intelligence (XAI)}
Explainable Artificial Intelligence refers to techniques that help make the decision processes of complex machine learning models understandable to humans. XAI methods are broadly divided into intrinsic and post-hoc approaches. Intrinsic models, such as decision trees, linear models, and rule-based learners, are transparent by design, although they may not achieve the predictive performance required for advanced fraud detection tasks. Post-hoc methods generate explanations for black-box models without modifying their internal mechanisms. Examples include LIME [5], which creates local linear approximations, and SHAP [6], which uses principles from cooperative game theory to measure feature contributions. Other methods include counterfactual explanations that show how small changes in the input could alter the prediction, as well as attention mechanisms that highlight influential input regions in deep models. While these techniques enhance interpretability, many remain difficult to scale in distributed Big Data environments.

\subsection{Risk Management and Fraud Detection}
Risk management and fraud detection are the top applications of Big Data analytics. The daily interactions of financial institutions, insurers, and public agencies in billions need to be evaluated for exposure to risk or fraudulent activity. Big Data also allows real-time scoring of risk exposure, dynamic exposure analysis, and the integration of internal and external signals to produce well-rounded (and often better revised) risk profiles in risk management. High-performance analytics systems enable organizations to address credit risk, operational risk, and market fluctuations more promptly and effectively. Big Data facilitates scalable anomaly detection, behavioral modeling, device fingerprinting, and graph-based analysis to detect fraud rings and coordinated schemes.[12] With the emergence of machine learning models, we are able to identify patterns that evolve over time and hence can develop adaptive fraud prevention systems. Recent developments that incorporate Explainable AI (XAI) and fraud analytics are targeting transparency and regulatory needs [7]. Some tools, SHAP, LIME, counterfactual explanations, and structural graph explanations are used in black-box fraud models to be interpretable by the analyst and the external parties, including auditors and end users. While XAI increases the computational burden, it also increases trust, helps satisfy compliance obligations, and improves investigative decisions.[14] With the integration of Big Data and XAI, we could achieve faster, more accurate, and more accountable fraud detection across a massive digital footprint.

\begin{table}[!t]
\caption{Comparison of Common XAI Techniques}
\centering
\begin{tabular}{|p{1.8cm}|p{2cm}|p{3.5cm}|}
\hline
\textbf{Method} & \textbf{Type} & \textbf{Strengths / Limitations} \\
\hline
LIME & Local & Easy to implement; provides local explanations but may be unstable across runs. \\
\hline
SHAP & Local/Global & Strong theoretical foundation; computationally expensive for large datasets. \\
\hline
Counterfactuals & Instance-based & Produces human-friendly explanations; difficult for high-dimensional inputs. \\
\hline
Attention & Intrinsic & Highlights influential features in deep models; not always guaranteed to reflect true model reasoning. \\
\hline
\end{tabular}
\label{table_xai}
\end{table}

Common post-hoc methods include LIME (Local Interpretable Model-Agnostic Explanations), SHAP (Shapley Additive Explanations), counterfactual explanations, attention mechanisms in deep networks, and feature importance visualization for ensemble models. While these methods improve interpretability, many struggle to scale efficiently within distributed Big Data environments such as Hadoop, Spark, or cloud-native architectures.

\subsection{Existing Integrations of XAI and Big Data}
Early research has attempted to combine explainability with large-scale analytics for domains such as credit risk scoring, money laundering detection, and cybersecurity intrusion monitoring. Numerous studies show that XAI improves user trust, supports regulatory compliance, and enhances the interpretability of automated risk decisions. For example, SHAP has been applied to justify credit approval decisions, and LIME has been used to identify anomalies in network traffic patterns.

Despite these advancements, significant limitations remain. Existing XAI methods often fail to scale across distributed infrastructures such as Hadoop, Spark, or real-time streaming platforms. Fraud detection systems must produce explanations within milliseconds, but most post-hoc models introduce substantial computational overhead. Additionally, explainability for graph-based models and streaming data remains relatively underdeveloped. Another challenge is the absence of standardized explanation formats that auditors, regulators, and end users can consistently interpret. These gaps highlight the need for more robust integration of XAI within Big Data analytics pipelines.

\section{Problem Statement and Research Gap}

Machine learning models are widely used today for fraud detection and risk assessment, but many of these systems still function as black boxes [5]. Complex models such as deep neural networks, boosted ensembles, and graph-based fraud detectors can make highly accurate predictions, yet they do not clearly show how those predictions are generated. This lack of transparency is a major concern in high-stakes settings where organizations must justify automated decisions to auditors, regulators, and customers.

Regulations such as the General Data Protection Regulation (GDPR), the EU AI Act, and U.S. financial guidelines now require automated systems to provide clear explanations [8], maintain audit trails, and demonstrate fairness. However, many existing Big Data fraud detection pipelines struggle to meet these expectations. These systems operate across distributed environments, process extremely large and fast-moving datasets, and must produce decisions in real time. Adding explainability to this workflow often slows down the system or introduces significant computational cost.

Current XAI techniques also have limitations [6]. Most post-hoc explanation methods require heavy computation and are not well suited for high-velocity streaming data. Distributed computing environments such as Hadoop, Spark, and cloud-native platforms further complicate the process of generating explanations at scale. In addition, graph-based fraud detection methods, which are essential for identifying fraud networks, still lack strong and scalable explanation tools.

Combined, these difficulties point out a clear research gap [7]. A single framework that would be able to deliver real-time, scalable, and reliable explanations within Big Data fraud detection pipelines does not exist. It is vital to confront this gap to be able to create trustworthy and compliant systems for automated decision-making.

\subsection{XAI Challenges in Big Data Environments}

Explainable AI encounters several difficulties when applied to Big Data systems. Many XAI techniques such as LIME and SHAP were originally developed for small or medium-sized datasets [5][6]. In such cases, when these techniques are applied to large-scale or high-speed data streams, they frequently turn out to be too slow or too costly from a computational point of view for practical use. Take, for instance, a fraud detection system that may be operating 1 million transactions per second, hence, issuing an account for each prediction becomes so resource-consuming that it is practically infeasible. Therefore, companies should trade off between precision, interpretability, and reaction time to make sure that the explanations do not interfere with decision-making in real time. 

\subsection{Gap in Scalable XAI for Big Data Pipelines}

While Big Data analytics has advanced significantly, explanation tools have not progressed at the same rate. Most XAI research focuses on small, static datasets [7], whereas real-world fraud detection relies on continuous multimodal data streams. Current methods frequently fail to produce explanations that remain consistent across numerical transactions, log data, sensor inputs, or graph structures. This creates a gap between the strong predictive performance of modern fraud detection systems and the need for transparent, auditable decisions. Addressing this gap is essential for building reliable and responsible AI systems for mission-critical applications.

\section{The Proposed REXAI-FD Integrated Framework}

\subsection{Introduction and Motivation}

The integration of Explainable AI (XAI) into large-scale risk management systems presents a formidable challenge, often creating a trilemma between model accuracy, operational scalability, and decision transparency. Contemporary XAI techniques, while powerful in isolation, frequently introduce prohibitive computational overhead and lack the standardization required for seamless deployment in distributed, real-time data pipelines [9][10]. This gap is particularly evident in financial fraud detection, where high-stakes decisions demand both rapid response and clear justifications.

Existing literature offers valuable but fragmented solutions. Studies have demonstrated the efficacy of distributed architectures for IoT security [9], the application of XAI for transaction monitoring [11], and the novel use of large language models (LLMs) for feature enrichment in anomaly detection [10]. However, a holistic framework that unifies these advances into a cohesive, enterprise-grade system remains elusive.

To address this, we propose the REXAI-FD framework (Real-time Explainable AI for Fraud Detection). Our contribution is a unified architecture that harmonizes three critical aspects: it leverages semantic intelligence from LLMs for richer feature representation, incorporates an adaptive explanation layer that dynamically matches explanation depth to operational context, and is built upon a cloud-native foundation for inherent scalability and resilience. This integrated approach moves beyond simply attaching explainability to a model, and instead bakes it directly into the fabric of a high-performance risk detection platform.

\subsection{Overall Architecture}

\begin{figure}[htbp]
    \centering
    \includegraphics[width=0.5\textwidth]{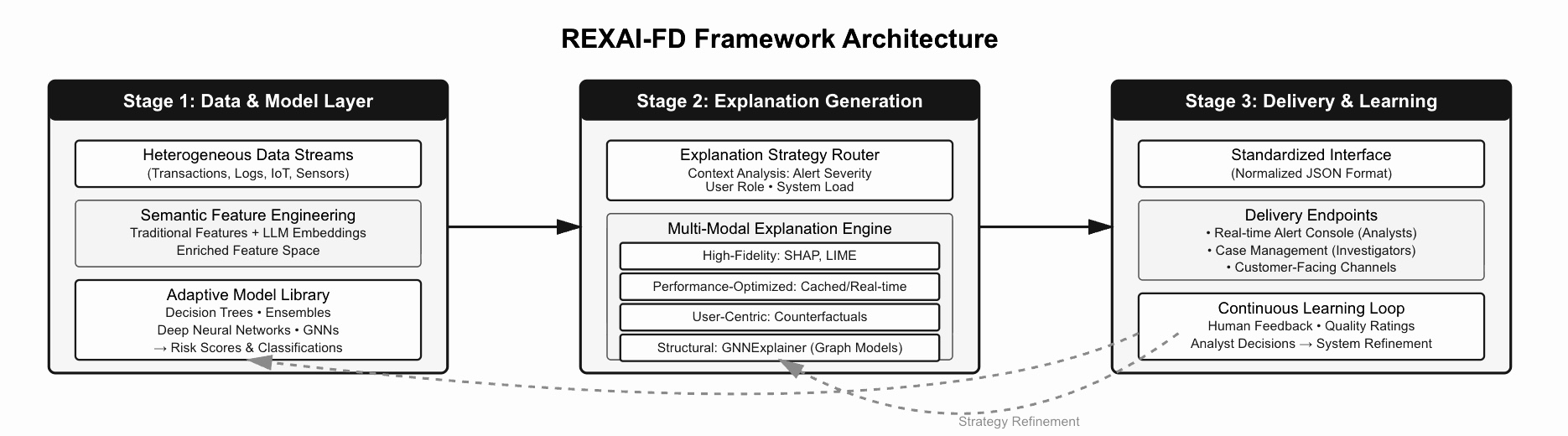}
    \caption{}
    \label{fig:framework}
\end{figure}

The REXAI-FD framework is conceived as a pipeline of intelligent, interconnected services rather than a monolithic application. Its design philosophy centers on providing the right explanation to the right user, at the right time, without compromising system performance. As illustrated in Figure~\ref{fig:framework}, the architecture flows through three primary stages:

\begin{enumerate}
    \item Data Ingestion and Multi-Model Inference: Heterogeneous data streams are processed through a feature engineering pipeline that combines traditional structured data with dense vector representations from LLM embeddings. This enriched feature set is then evaluated by a suite of machine learning models, selected to balance performance and interpretability based on the specific risk scenario.
    \item Context-Aware Explanation Generation: The output from the model layer, a risk score and classification triggers a dynamic explanation process. A central Explanation Strategy Router acts as an intelligent dispatcher, analyzing contextual cues such as the severity of the alert, the user role requesting insight, and system load to select the most appropriate explanation methodology from a suite of specialized explainers.
    \item Actionable Delivery and Continuous Learning: Generated explanations are formatted into a consistent schema and delivered to tailored user interfaces. Crucially, this layer closes the loop by capturing feedback from human analysts, using their validated judgments and quality ratings to continuously refine both the detection models and the explanation strategies themselves.
\end{enumerate}

This layered, event-driven design ensures that the system can meet the low-latency demands of real-time fraud prevention while also supporting the deep, retrospective analysis required for forensic investigations and model auditing.

\subsection{Core Components}

\subsubsection{Data and Model Layer}

This layer is responsible for transforming raw data into actionable intelligence. Its hybrid design allows it to navigate the trade-off between predictive power and explainability.

\begin{itemize}
    \item Semantic Feature Engineering: Moving beyond traditional feature extraction, this component integrates pre-trained LLM embedding models [10]. By converting textual data such as transaction narratives or system logs into high-dimensional vectors, it captures nuanced semantic relationships that are often invisible to frequency-based methods. This process enriches the feature space, allowing models to detect more sophisticated fraud patterns and providing a richer substrate for subsequent explanation.
    \item Adaptive Model Library: The framework maintains a repository of machine learning models, ranging from highly interpretable classifiers like Decision Trees to complex ensembles and deep learning networks. The selection of a model for a given task is not static; it can be configured based on the prevailing requirements for accuracy, speed, or regulatory-mandated transparency, ensuring operational flexibility.
\end{itemize}

\subsubsection{Explanation Generation Layer}

Serving as the intellectual core of the framework, this layer transforms model outputs into understandable rationales. Its strength lies in its ability to avoid a one size fits all approach.

\begin{itemize}
    \item Explanation Strategy Router: This lightweight service makes critical routing decisions in real-time. For a low risk transaction flagged for a simple log, it might retrieve a precomputed feature importance score. In contrast, a high-value transaction under manual review would be routed to a more computationally intensive explainer for a deep-dive analysis. This dynamic allocation optimizes resource usage and user experience simultaneously.
    \item Multi-Modal Explanation Engine: The router draws upon a suite of specialized explainers:
    \begin{itemize}
        \item High-Fidelity Explainers (e.g., SHAP, LIME): Used for deep-dive analysis, these methods provide detailed, feature-level attribution for individual predictions, ideal for auditor reports and model diagnostics [10][11].
        \item Performance-Optimized Explainers: For real-time dashboards, these provide instant, approximate explanations based on model-internal metrics or cached values, ensuring minimal latency [11].
        \item User-Centric Explainers (e.g., Counterfactuals): These generators create natural language "what-if" scenarios (e.g., “This payment was blocked due to an unusual device. Had it been made from your registered phone, it would have been approved”) to communicate decisions effectively to end-users [9].
        \item Structural Explainers (e.g., GNNExplainer): For graph-based models that identify organized fraud rings, these tools visualize the subgraph structures and relational pathways that led to a suspicious classification [9].
    \end{itemize}
\end{itemize}

\subsubsection{Explanation Delivery and Human-in-the-Loop Layer}

The ultimate value of an explanation lies in its consumption and utility. This layer ensures insights are delivered effectively and used to foster continuous improvement.

\begin{itemize}
    \item Standardized Interface and Routing: All explanations are normalized into a standard JSON format, ensuring consistency across the platform. They are then routed to the appropriate endpoints: a real-time alert console for security analysts, a comprehensive case management system for investigators, or customer-facing communication channels.
    \item Continuous Learning Loop: The framework treats human feedback as a first-class data source. When an analyst overrules a model's decision or provides a quality rating on an explanation, this signal is fed back into the system. Confirmed labels are used to retrain and improve the detection models, while explanation feedback helps the Strategy Router learn which explanation modes are most effective in different contexts, creating a self-improving system.
\end{itemize}

\subsection{Implementation Considerations}
For REXAI-FD to be deployed in a production environment, a modern, cloud-native technology stack is essential. The framework should be implemented as a collection of loosely coupled microservices, containerized using Docker and orchestrated with Kubernetes to ensure resilience, auto-scaling, and operational manageability [9]. To address the computational demands of techniques like KernelSHAP, distributed processing frameworks such as Apache Spark can parallelize workloads across a cluster, making intensive explanation generation feasible on large datasets [11]. For real-time processing, streaming technologies like Apache Kafka enable asynchronous handling of transaction data, high-risk events can be published to dedicated topics, decoupling detection workflows from explanation services, and preventing bottlenecks [11]. Finally, to optimize the integration of LLMs, balancing cost and latency, a caching strategy for embedding vectors is recommended. By pre-computing and storing embeddings for frequently occurring text patterns, external API calls can be significantly reduced, improving both response times and system efficiency [10].

\subsection{Necessity of Interpretability for Risk-Sensitive Domains}
Regulatory, ethical, and operational pressures are pushing complex sectors that are subject to high levels of risk like banking, insurance, and healthcare to have interpretable machine learning. Big Data enables risk scoring and anomaly detection with high precision, but black-box models which are not visible introduce other levels of uncertainty to an auditor, regulator, and end user. Explainable AI solves this by detailing how a model would identify a customer as high risk, deny a loan application, or flag a transaction as fraudulent. Such clarity not only helps to conform with the evolving AI governance structures but also permits human intervention to verify whether the alerts are due to a legitimate reason or biased or unstable model behaviour. Thus, incorporation of XAI is becoming the underlying necessity to responsibly deploy Big Data analytics in fields where transparency is not a benefit, it's a prerequisite.

\section{Discussion}

This section discusses the broader implications of integrating explainable artificial intelligence (XAI) into big data analytics for fraud detection and risk management. The findings from the literature review and analysis highlight the tension between accuracy, scalability, and interpretability in modern automated decision systems.

\subsection{Human-in-the-Loop Decision Making Enhanced by XAI}
The implementation of Explainable AI is a breakthrough in human-in-the-loop decision methods of Big Data analytics. Rather than simply replacing human judgment, XAI supports analytical capability in order to work more efficiently with ML models—giving transparent reasoning for every prediction or alert. This partnership has great value in fraud investigations, where analysts need to decide whether suspicious behavior is really malicious or just unusual but legitimate. XAI improves investigation speed and accuracy, reduces false positives, and establishes user trust in AI-based processes by illuminating the most influential features or patterns behind a model’s decision. In the end, XAI turns the complexity of analytics systems from opaque automation software into collaborative decision-support systems.

\subsection{Theoretical Implications}

The integration of XAI into large-scale analytics presents new theoretical considerations. First, it suggests that explainability must be treated as a core system requirement and not as an “add-on” to existing models. Many traditional fraud detection algorithms, especially deep neural networks and graph-based models, offer strong predictive performance but lack interpretability. XAI introduces mechanisms that bridge this gap by creating structured, human-understandable explanations. However, the literature indicates unresolved questions regarding the fidelity of these explanations, their stability under distributed computation, and their suitability for high-dimensional, high-velocity data environments.

\subsection{Practical Implications}

From an operational perspective, XAI improves organizational trust and supports regulatory compliance by enabling model transparency in financial decision systems. Analysts gain insight into which features influence fraud predictions, facilitating better case investigations and faster resolution times. Institutions subject to audit requirements benefit from standardized explanations that document model behavior. Real-time systems, such as payment monitoring platforms and cybersecurity intrusion detection, stand to gain from lightweight explanation mechanisms that balance latency with interpretability.

\subsection{Limitations}

Despite these benefits, several limitations persist. Current XAI methods introduce substantial computational overhead, making them difficult to deploy in distributed or streaming environments. Real-time fraud detection systems require millisecond-level performance, whereas many XAI methods operate at seconds or minutes per explanation. Existing techniques also struggle to explain predictions for graph neural networks, temporal models, or large ensemble systems. Additionally, there is limited standardization in explanation formats, reducing their usefulness in regulatory audits and cross-institutional reporting.

\section{Challenges and Open Research Questions}

Although explainability is gaining traction, several open challenges must be addressed before XAI can be fully integrated into large-scale risk management systems.

\subsection{Scalability of XAI for Distributed Big Data Systems}

Most XAI algorithms, such as SHAP and LIME, were designed for single-machine environments and do not scale naturally to platforms such as Hadoop, Spark, or cloud-native streaming systems. Research is needed to develop parallelized or approximate explainability methods that retain acceptable fidelity while operating at cluster scale.

\subsection{Balancing Privacy, Explainability, and Big Data Scale}
One of the big problems that the XAI driven Big Data analytics has had to address is the privacy, interpretability and system scalability trade-off. As a result, most explanation techniques continue to rely on raw features or sensitive attributes which can violate widely-accepted data minimization or differential privacy practice in financial and/or healthcare contexts. At the same time, simplifying the data for privacy purposes could result in the explanations being less reliable and therefore less useful to the analyst. Meaningful interpretability without compromising confidentiality necessitates investigations of privacy-preserving XAI techniques (e.g. federated explanations, encrypted feature attribution, aggregated surrogate models).[15] Developing such approaches is very important as organizations strive to identify information from extremely sensitive, huge datasets.

\subsection{Real-Time Explanations for Streaming Fraud Detection}

High-frequency environments, such as credit card authorizations and digital banking transactions, require explanations to be generated in milliseconds. Achieving real-time explainability without sacrificing detail or accuracy remains an unsolved challenge. Techniques that pre-compute explanation templates or leverage model-internal attributions are promising but underexplored.

\begin{figure}[h!]
\centering
\begin{tikzpicture}[
node distance=1.4cm,
process/.style={rectangle, draw, rounded corners, minimum width=3.2cm, minimum height=1cm, align=center},
arrow/.style={-stealth, thick}
]

\node[process] (data) {Streaming Transaction Data};
\node[process, below=of data] (preprocess) {Preprocessing \\ Normalization, Feature Extraction};
\node[process, below=of preprocess] (model) {ML Model \\ (GNN, LSTM, XGBoost)};
\node[process, below=of model] (xai) {XAI Engine \\ (SHAP, LIME, Counterfactuals)};
\node[process, below=of xai] (decision) {Human + AI Decision};
\node[process, below=of decision] (action) {Approve / Block / Investigate};

\draw[arrow] (data) -- (preprocess);
\draw[arrow] (preprocess) -- (model);
\draw[arrow] (model) -- (xai);
\draw[arrow] (xai) -- (decision);
\draw[arrow] (decision) -- (action);

\end{tikzpicture}
\caption{Real-time fraud detection pipeline integrating Explainable AI.}
\end{figure}
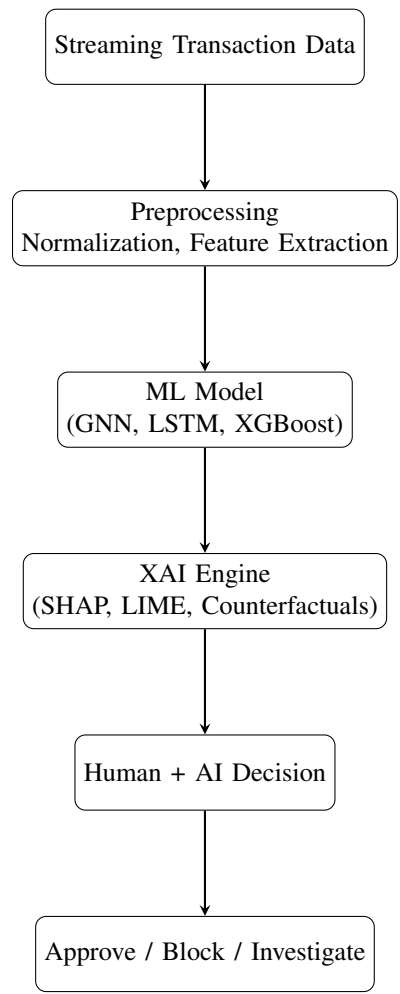

\subsection{Explainability for Graph and Deep Temporal Models}

Fraud rings and coordinated attacks are often detected using graph neural networks or temporal models, yet explanation frameworks for these architectures remain limited. Research must address how to visualize subgraph importance, temporal dependencies, and relational reasoning in a way that domain experts can understand.

\subsection{Human-Centered and Regulatory-Aligned Explanations}

Different stakeholders require different forms of explanation. Fraud analysts need feature-level attributions, auditors need stable and reproducible explanations, and customers require simple natural-language justifications. Creating a unified explanation schema that meets these diverse needs while remaining compliant with regulations such as GDPR and the EU AI Act is an important research question.

\subsection{Standardization and Benchmarking}

There is no standard benchmark for evaluating XAI performance in financial fraud detection. Future work should propose publicly accessible datasets, standardized explanation taxonomies, and objective metrics for evaluating explanation quality and regulatory adequacy.

\section{Conclusion}

This paper examined the role of explainable artificial intelligence in enhancing transparency, accountability, and trust in big data–driven fraud detection and risk management systems. While machine learning models offer remarkable accuracy and adaptability, their opaque nature presents challenges for regulatory compliance and operational oversight. Explainable AI provides a promising pathway toward addressing these issues by making complex model behavior interpretable without sacrificing performance.

The analysis revealed that existing XAI techniques offer valuable insights but face challenges related to scalability, real-time execution, and applicability to advanced architectures such as graph neural networks. A comprehensive integration of XAI into big data pipelines requires further research on distributed explainability, human-centered explanation frameworks, and standardized formats suitable for regulatory audits.

Overall, this work highlights the importance of combining high-performance analytics with transparent, interpretable decision processes. Continued advancements in XAI will be critical to ensuring that automated fraud detection systems remain both effective and trustworthy as data volumes and model complexities continue to grow.

\end{document}